\documentclass[conference]{IEEEtran}
\IEEEoverridecommandlockouts

\usepackage{amsmath,amssymb,amsfonts}
\usepackage{algorithm,algorithmic}
\usepackage{graphicx}
\usepackage{textcomp}
\usepackage{xcolor}
\usepackage{booktabs}
\usepackage{multirow}
\usepackage[normalem]{ulem}
\useunder{\uline}{\ul}{}

\usepackage[style=ieee, 
            citestyle=numeric-comp,
            maxnames=2, 
            minnames=1, 
            doi=false,
            isbn=false,
            url=false,
            date=year,
            ]{biblatex}
\begin{document}

\title{Texo: Formula Recognition within 20M Parameters\\
\thanks{We thank Mathieu Fontaine and Anthony Larcher for reviewing our paper and we thank Telecom Paris for providing GPU resources.}
}


\author{\IEEEauthorblockN{Sicheng Mao}
\IEEEauthorblockA{\textit{LTCI, Télécom Paris, Institut Polytechnique de Paris} \\
\textit{Paris, France} \\
sicheng.mao@telecom-paris.fr}
}

\maketitle

\begin{abstract} 
In this paper we present Texo, a minimalist yet high-performance formula recognition model that contains only 20 million parameters. By attentive design, distillation and transfer of the vocabulary and the tokenizer, Texo achieves comparable performance to state-of-the-art models such as UniMERNet-T and PPFormulaNet-S, while reducing the model size by 80\% and 65\%, respectively. This enables real-time inference on consumer-grade hardware and even in-browser deployment. We also developed a web application to demonstrate the model capabilities and facilitate its usage for end users. 
\end{abstract}


\section{Introduction}
\label{sec:intro}
\begin{figure}[t]
    \centering
    \includegraphics[width=0.4\linewidth]{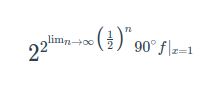}
    \includegraphics[width=0.8\linewidth]{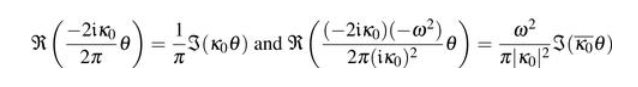}
    \includegraphics[width=0.8\linewidth]{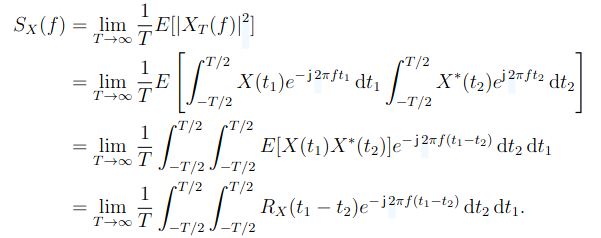}
    \includegraphics[width=0.8\linewidth]{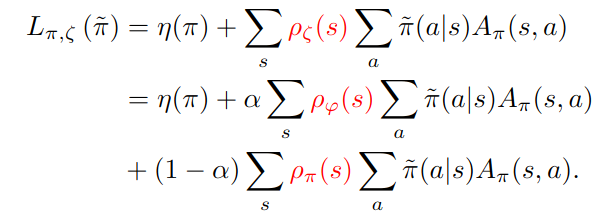}
    \caption{Example images that Texo is able to recognize.}\footnotesize{Check https://texocr.netlify.app to find the recognition results.}  
    \label{fig:example-images}
\end{figure}

Formula recognition, also known as Mathematical Expression Recognition (MER), constitutes a critical component of document analysis. By converting formula images into structured LaTeX code (as in Fig.~\ref{fig:example-images}), MER facilitates note-taking and academic writing for scientists, teachers, and students \cite{mathpix}. Beyond this direct usage, MER also plays an increasingly important role in the era of Large Language Models (LLMs). Modern LLMs rely heavily on high-quality training data, a substantial portion of which is extracted from academic documents~\cite{Liu2024LLMs}. Therefore, the accurate and efficient extraction of mathematical expressions from these documents is essential for data preprocessing in the LLM training pipeline.

The research on the MER task started from the traditional machine learning era. \cite{anderson1967syntax} pioneered MER in irregular documents, followed by an integrated character segmentation and layout grammar system \cite{Miller_Viola_1998}, an online recognition error correction system \cite{Chan_Yeung_1999} and a mathematical document optical character recognition (OCR) system \cite{suzuki2003infty}. However, these methods were limited by handcrafted features.

In the deep learning era, convolutional neural networks (CNNs) \cite{krizhevsky2012imagenet,simonyan2015very} revolutionized MER as a powerful image feature extractor. Key advances include encoder-decoder models with coarse-to-fine attention \cite{deng2017image}, grammar and segmentation self-learning models \cite{zhang2017watch}, handwritten MER via paired adversarial learning \cite{wu2020handwritten}, tree-structured decoders for complex markups \cite{zhang2020tree}, bidirectional learning-enhanced architectures \cite{zhao2021handwritten, bian2022handwritten}, weakly supervised counting modules \cite{li2022counting} and data augmentation strategies \cite{le2019pattern,li2020improving} to enhance MER performance. 

Recent progress in Transformer models \cite{vaswani2017attention} and vision-language models (VLMs) \cite{zhu2023minigpt,liu2024improved,chen2024far} has driven researchers to explore the more general document parsing task, which naturally includes MER as a sub-task. End-to-end models such as Donut \cite{kim2022ocr}, Nougat \cite{blecher2023nougat} and GOT-OCR-2.0 \cite{wei2024general} have advanced academic document understanding in this trend, but their general-purpose design overlooks MER-specific characteristics limiting their performance on this task. On the other hand, MER-specific models such as Pix2tex \cite{pix2tex2022} and Texify \cite{texify2023} perform well on simple expressions but still struggle with complex or noisy expressions. To address these issues, \cite{wang2024unimernet} open-sourced the curated UniMER dataset and the UniMERNet models, achieving the current state-of-the-art (SOTA) among the open-source models. Based on this, \cite{liu2025PPFormulaNet} trained PPFormulaNet over an in-house dataset and further optimized its performance with an efficiency trade-off.

However, the above open-source SOTA MER models are still too heavy to run on personal computers or other edge devices; for example, the size of the UniMERNet model series ranges from 107M (Tiny) to 325M (Base) and the size of the PPFormulaNet series ranges from 58M (Small) to 181M (Large). In comparison, the size of GOT-OCR-2.0 is 560M. The parameter efficiency is crucial for such a task-specific model; otherwise, a general purpose VLM would be more favorable to deploy in practice. 

\newpage

In this paper, we address the parameter efficiency issue of the MER models. Our contributions are as follows:

\begin{itemize}
    \item We propose Texo, a minimalist formula recognition model with performance comparable to the SOTA models, especially by curating a domain-specific vocabulary and tokenizer and preserving as much knowledge as possible through vocabulary distillation and transfer from the PPFormulaNet-S. We demonstrate the possibility of challenging large models on MER-specific task with only open source data, open weight models and consumer-grade hardware.   
    \item We serve our model in the browser, providing an accurate, fast, free, private and convenient formula recognition solution. 
\end{itemize}

\section{Method}
\subsection{Model Design}
To achieve minimal model size while still preserving good performance, we conduct a comprehensive technical selection over previous formula recognition models mentioned in Section \ref{sec:intro}. The parameter counts of some representative formula recognition models are listed in Table \ref{tab:model-stats}.

\begin{figure}[t]
    \centering
    \includegraphics[width=0.7\linewidth]{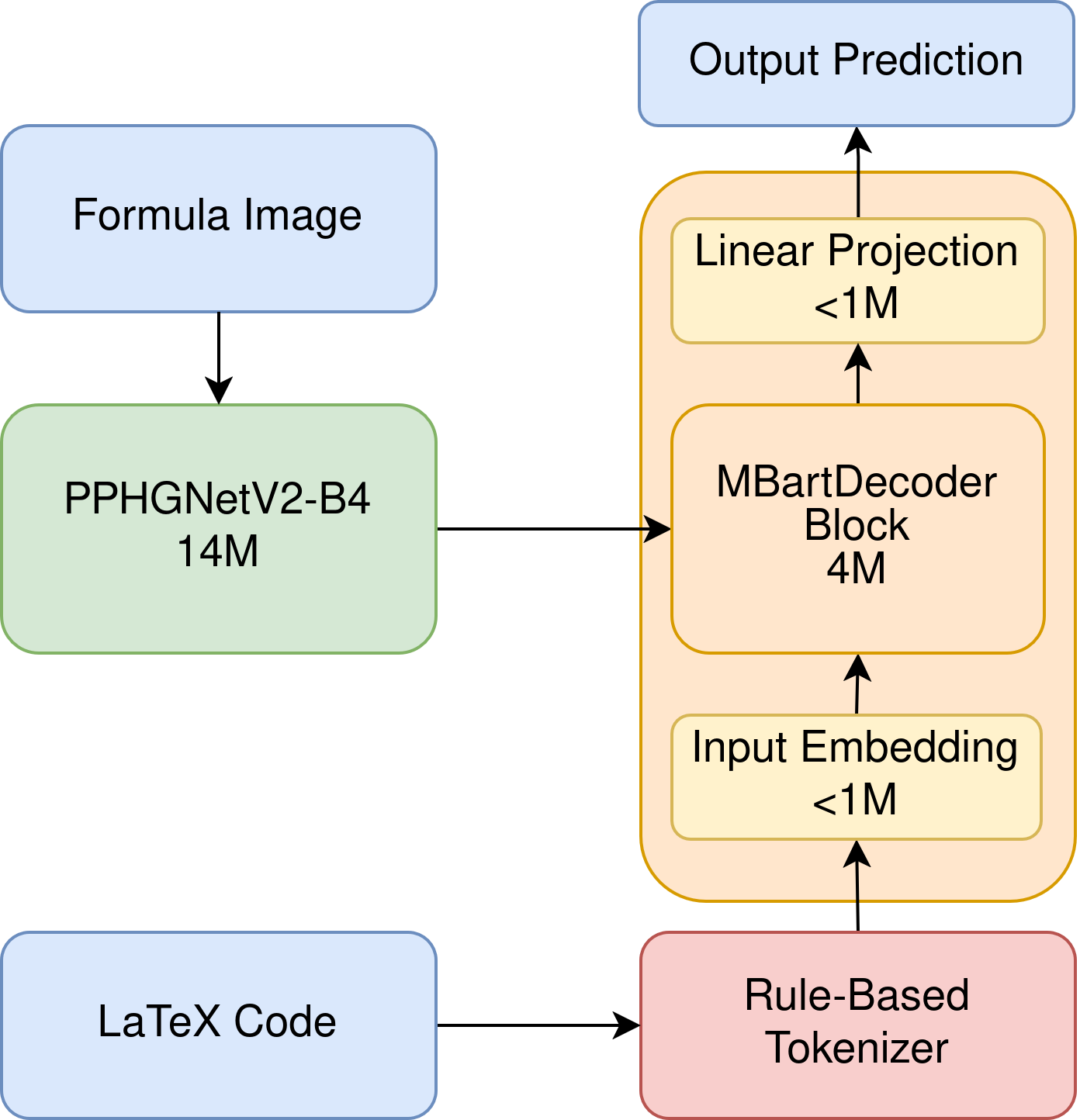}
    \caption{Texo model architecture in the training process}
    \label{fig:texo-model}
\end{figure}

Specifically, Texo inherits the model architecture mainly from PPFormulaNet-S, which consists of an image encoder and a text decoder, see Fig. \ref{fig:texo-model}. The image encoder is implemented by HGNetV2-B4, a lightweight modern CNN that serves as the backbone of RT-DETR, which is comparable to ViT on image classification and object detection tasks \cite{lv2023detrs,lv2024rtdetrv2improvedbaselinebagoffreebies}. The text decoder is a 2-layer MBart Transformer decoder \cite{liu2020multilingualdenoisingpretrainingneural} with a hidden dimension equal to 384 and a context length of 1024. The encoded image hidden features are cross-attended by the text decoder, which outputs a sequence of log-probabilities for cross-entropy loss minimization training.  

\begin{table}[h]
\centering
\caption{Parameter counts of formula recognition models}
\label{tab:model-stats}
\begin{tabular}{llll}
\hline
Model       & GOT-OCR2.0 & UniMERNet-T & PPFormulaNet-S \\ \hline
Encoder     & ViT 95M    & SwinT  25M  & HGNetV2 14M    \\
Decoder     & Qwen2 463M & MBart 81M   & MBart 44M      \\
Vocab. size & 152K       & 50K         & 50K            \\
Hidden. dim. & 1024       & 512         & 384            \\
Embedding   & 155M       & 51M         & 38M\\
Total       & 560M       & 107M        & 58M            \\ \hline
\end{tabular}
\end{table}

\subsection{Model compression by vocabulary distillation and transfer}

Previous works \cite{wang2024unimernet, liu2025PPFormulaNet} primarily reduce model size by decreasing the hidden dimensionality or the number of layers of the Transformer decoder. However, a straightforward parameter count reveals that the dominant parameter overhead arises from the decoder’s input and output embedding layers, whose sizes are proportional to the vocabulary size, as shown in Table \ref{tab:model-stats}. For almost all of the formula recognition models mentioned in Section \ref{sec:intro}, the vocabulary is inherited from their text decoders built for general natural language tasks. This poses two problems: 1) the vocabulary built for natural language corpus is much larger than the one that formula corpus requires, resulting in a waste of tokens and hence parameters of embedding layers; 2) the common Byte Pair Encoding (BPE) tokenizer equipped with these models is trained in a heuristic manner, which may break the meaningful token into less significant subwords, e.g. ``$\backslash$leftarrow" is split into ``$\backslash$left" and ``arrow".  

As the LaTeX formula is more syntactically rigorous and semantically limited compared to natural language, we first distill the vocabulary to greatly reduce the number of tokens. Specifically, 
we refer to KaTeX \cite{katex}, an open-source LaTeX parser, to obtain all common macros and build a tokenizer in a rule-based manner since the LaTeX macros carry complete semantics, while the BPE tokenizer may cut them into subwords, leading to longer sequence lengths and looser semantic dependencies. To adapt the model to the new tokenizer, we adapt the vocabulary transfer heuristics proposed in \cite{geeFastVocabularyTransfer2022, mosinFinetuningTransformersVocabulary2023}. Specifically, we remove the whitespace from the tokenizer as it has no semantic significance. This drastically reduces the sequence length of the training data and promotes more effective modeling for complex formulae. The pseudo code is given in Algorithm \ref{alg:vocab-transfer}. 

The vocabulary distillation and transfer reduce the vocabulary size from 50,000 to 687; hence, the embedding parameters decrease from 38 million to less than 1 million. As a result, the distilled model contains only 20 million parameters in total, which is approximately 35\% of the size of PPFormulaNet-S. 

\subsection{In-browser deployment}
To not only demonstrate the inference efficiency but also facilitate the usage, we deploy our model in a purely front-end manner without any back-end API calls. To build the web application, we export the best Texo checkpoint to ONNX format, serve it using the Transformers.js framework, and construct the user interface using Vue, Nuxt and NuxtUI. We use the Web Worker to separate the inference from the UI rendering for a better user experience. We also embed functionalities such as conversion to Typst and MathML, as well as a \textit{what you see is what you get} (WYSIWYG) editor.    

For general users, the Texo web application provides a convenient and confidential way to run the model on their local computational resources. This not only eliminates the troublesome environment configuration found in common open-source solutions but also avoids the risks of personal data leaks through API calls in commercial solutions.

\begin{algorithm}[H]
\caption{Vocabulary Transfer}
\label{alg:vocab-transfer}
\begin{algorithmic}[1]
\STATE \textbf{Input:} base tokenizer $T$ with its vocabulary $V$, target tokenizer $T'$ with its vocabulary $V'$ and base model decoder $D$
\STATE \textbf{Output:} Vocabulary-transferred decoder $D'$

\STATE Initialize token mapping dictionary $\mathcal{M} \gets \emptyset$
\FOR{each token $t \in V'$}
    \IF{$t \in V$ \AND \textvisiblespace $+t \in V$}
        \STATE $ids \gets T.\text{encode}$(\textvisiblespace$+t)$
    \ELSE
        \STATE $ids \gets T.\text{encode}(t)$ \\\COMMENT{token in $V'$ can be always tokenized by $T$}
    \ENDIF
    \STATE $\mathcal{M}[T'.\text{token\_to\_id}(t)] \gets ids$
\ENDFOR

\STATE $E_{in} \gets M.\text{input\_embeddings}$
\STATE $d \gets E_i.$\text{hidden\_dim}
\STATE $E_{in}' \gets \text{nn.Embedding}(|V'|, d)$
\FOR{each $(i, ids) \in \mathcal{M}$}
    \STATE $E_{in}'[i] \gets \text{torch.mean}(E_{in}[\text{sorted}(ids)])$
\ENDFOR
\STATE $D'.\text{set\_input\_embeddings}(E_{in}')$

\STATE $E_{out} \gets D.\text{output\_embeddings}$ 
\STATE $E_{out}' \gets \text{nn.Linear}(d, |V'|)$
\FOR{each $(i, ids) \in \mathcal{M}$}
    \STATE $E_{out}'[i] \gets \text{torch.mean}(E_{out}[\text{sorted}(ids)])$
\ENDFOR
\STATE $D'.\text{set\_output\_embeddings}(E_{out}')$

\RETURN $D'$
\end{algorithmic}
\end{algorithm}

\begin{table*}
\centering
\caption{The recognition performance of different models on the UniMER-Test dataset.}
\label{tab:scores}
\begin{tabular}{@{}ccccccccccc@{}}
\toprule
\multirow{2}{*}{Model} &
  \multicolumn{1}{c|}{\multirow{2}{*}{Params.}} &
  \multicolumn{4}{c|}{CDM ↑} &
  \multicolumn{4}{c|}{Token length ↓} &
  \multirow{2}{*}{\begin{tabular}[c]{@{}c@{}}GPU inference time\\ per sample(ms) ↓\end{tabular}} \\
 &
  \multicolumn{1}{c|}{} &
  SPE &
  CPE &
  SCE &
  \multicolumn{1}{c|}{HWE} &
  SPE &
  CPE &
  SCE &
  \multicolumn{1}{c|}{HWE} &
   \\ \midrule
Mathpix &
  - &
  0.973 &
  0.967 &
  0.924 &
  0.932 &
  - &
  - &
  - &
  - &
  - \\ \midrule
UniMERNet-T &
  107M &
  \textbf{0.991} &
  \textbf{0.949} &
  \textbf{0.938} &
  \textbf{0.933} &
  {\ul 114.82} &
  475.52 &
  {\ul 35.93} &
  {\ul 34.99} &
  2266.96 \\
PPFormulaNet-S &
  {\ul 57M} &
  0.949 &
  0.678 &
  0.856 &
  0.818 &
  117.10 &
  {\ul 461.16} &
  45.95 &
  41.44 &
  \textbf{217.60} \\
Texo &
  \textbf{20M} &
  {\ul 0.958} &
  {\ul 0.825} &
  {\ul 0.882} &
  {\ul 0.902} &
  \textbf{62.63} &
  \textbf{242.77} &
  \textbf{20.92} &
  \textbf{21.48} &
  {\ul 311.26} \\ \bottomrule
\end{tabular}
\end{table*}

\section{Experiment}

\subsection{Dataset}
We use the UniMER dataset \cite{wang2024unimernet} to train and evaluate our model. It consists of two splits: UniMER-1M, which contains over 1 million image-latex training pairs covering both simple and complex expressions in printed and handwritten styles; and UniMER-Test, which contains over 23,000 test pairs including 4 types of mathematical expressions: Simple Printed Expressions (SPE), Complex Printed Expressions (CPE), Screen-Captured Expressions (SCE), and Handwritten Expressions (HWE). 

We remove noise from the dataset such as redundant nested braces and synonymous LaTeX tokens based on the normalizer given in \cite{deng2017image}.

\subsection{Training}
We initialize our model with the pretrained weights from PPFormulaNet-S by converting from the original Paddle format to PyTorch, followed by the vocabulary transfer. 

We then train it on the UniMER-1M dataset.
The input formula images are first gray-scaled, white-margin cropped and resized to $384\times 384$ resolution. To further enhance recognition robustness, we employ data augmentation methods consistent with UniMERNet \cite{wang2024unimernet}, including mathematical morphological operations, affine transformations, Gaussian noise, brightness and contrast adjustments and weather perturbation. We use the AdamW optimizer for optimization with $\beta_1=0.9, \beta_2=0.999$ and a weight decay of 0.05. The batch size is set to 64 and the total number of training steps is set to $10^5$. The learning rate is set to $10^{-5}$ with a linear warm-up of 5,000 steps starting from $10^{-8}$ followed by a cosine annealing schedule. Training is conducted on a single A40 46GB GPU.\footnote{In comparison, the UniMERNet models were trained on 8$\times$A100 80GB GPUs \cite{wang2024unimernet} and the PPFormulanet models \cite{liu2025PPFormulaNet} were trained on 4 GPUs.} By reducing the batch size and enabling gradient accumulation, the training can be adapted to a consumer-grade GPU such as the RTX3090 24GB, as the parameters of the model and optimizer take only 230MB.

\subsection{Evaluation}
We perform evaluations on the UniMER-Test dataset.
We compare Texo with the two state-of-the-art models, namely UniMERNet-T and PPFormulaNet-S.
We use the Character Detection Matching (CDM) score implemented in \cite{wangImageTextTransforming2025} instead of traditional sequential metrics such as the BLEU score or Edit Distance, as it is more robust to textually variable but visually equivalent LaTeX expressions and has become the de-facto standard metric for formula recognition \cite{ouyang2024omnidocbench}. We also evaluate the output prediction token length and the inference speed of each model, averaged over the entire UniMER-Test. The inference time is measured on a single A40 46GB GPU with a batch size equal to 1. The evaluation results are reported in Table \ref{tab:scores}. For reference, we also list the CDM performance of Mathpix \cite{mathpix}, which is a commercial SOTA model.

We observe that although the Texo model contains the least number of parameters, it still achieves comparable recognition performance on formula images under different conditions. Particularly, it consistently outperforms its base model PPFormulaNet-S, since our vocabulary distillation method makes the classification head concentrate only on the useful tokens and reduces erroneous predictions. On the other hand, the curated tokenizer reduces nearly half of the output token length, which also contributes to the inference speed as the prediction would end faster. Speaking of the inference speed, our model is $7\times$ faster than UniMERNet-T, while it still falls a little behind PPFormulaNet-S which uses the multi-token parallel prediction technique with parallel steps equal to 3, hence $3\times$ faster than the normal prediction. However, this technique sacrifices accuracy for speed which we do not apply, considering the already fast inference speed thanks to our distillation method.

\section{Conclusion}
We propose Texo, a minimalist yet high-performance formula recognition model that contains only 20 million parameters. It achieves comparable
performance to state-of-the-art models such as UniMERNet-T
and PPFormulaNet-S, while reducing the model size by 80\% and 65\% respectively and accelerating the inference by 7$\times$ faster than UniMERNet-T. With careful model and tokenizer design, as well as vocabulary distillation and transfer techniques, we demonstrate the possibility of challenging the state-of-the-art large models on the MER-specific task with limited resources. We serve our model in the browser to demonstrate its efficiency. We open source the model and the training pipeline for pedagogical purposes. Following the recent boost of large vision language models for general document OCR tasks \cite{cui2026paddleocrvl15, wei2025deepseek, hunyuanocr2025}, a natural continuation of Texo would be to adapt the parameter efficient model to more general document OCR tasks as well.

\section{Reference}
\printbibliography[heading=none]
\end{document}